\documentclass[letterpaper, 10 pt, conference]{ieeeconf}  

\IEEEoverridecommandlockouts                              

\usepackage[hidelinks]{hyperref}
\usepackage{booktabs}
\usepackage[utf8]{inputenc}
\usepackage[T1]{fontenc}
\usepackage{amsmath,amssymb,amsfonts}
\usepackage{xcolor}
\usepackage{algorithm} 
\usepackage{algpseudocode} 
\usepackage{graphicx} 
\usepackage{caption} 
\usepackage{url}
\usepackage{flushend}

\title{\LARGE \bf
WheelArm-Sim: A Manipulation and Navigation Combined Multimodal Synthetic Data Generation Simulator for Unified Control in Assistive Robotics \vspace{-0.5em}
}

\author{Guangping Liu$^{1}$, 
Tipu Sultan$^{1}$,
Vittorio Di Giorgio$^{2}$,
Nick Hawkins$^{1}$,
Flavio Esposito$^{2}$,
Madi Babaiasl$^{1}$ 
\thanks{$^{1}$Guangping Liu, Tipu Sultan, Nick Hawkins, and Madi Babaiasl are with Aerospace and Mechanical Engineering Department, Saint Louis University, St. Louis, MO, 63103, United States. Dr. Madi Babaiasl is the corresponding author of this paper ({madi.babaiasl@slu.edu}).}%
\thanks{$^{2}$Flavio Esposito and Vittorio Di Giorgio are with Computer Science Department, Saint Louis University, St. Louis, MO, 63103, United States.} 
\thanks{Accepted to IEEE International Symposium on Medical Robotics (ISMR) 2026.}  }     

\begin{document}

\maketitle
\thispagestyle{empty}
\pagestyle{empty}


\begin{abstract}

Wheelchairs and robotic arms enhance independent living by assisting individuals with upper-body and mobility limitations in their activities of daily living (ADLs). Although recent advancements in assistive robotics have focused on Wheelchair-Mounted Robotic Arms (WMRAs) and wheelchairs separately, integrated and unified control of the combination using machine learning models remains largely underexplored. To fill this gap, we introduce the concept of WheelArm, an integrated cyber-physical system (CPS) that combines wheelchair and robotic arm controls. Data collection is the first step toward developing WheelArm models. In this paper, we present WheelArm-Sim, a simulation framework developed in Isaac Sim for synthetic data collection. We evaluate its capability by collecting a manipulation and navigation combined multimodal dataset, comprising 13 tasks, 232 trajectories, and 67,783 samples. To demonstrate the potential of the WheelArm dataset, we implement a baseline model for action prediction in the mustard-picking task. The results illustrate that data collected from WheelArm-Sim is feasible for a data-driven machine learning model for integrated control. The data, codes, and video of the paper can be accessed at https://github.com/madibabaiasl/WheelArmSimISMR2026. 

\end{abstract}
\vspace{-0.6em}

\section{Introduction}
\label{sec:introduction}Wheelchairs and robotic arms are critical assistive technologies for individuals with mobility or motor impairments. The World Health Organization (WHO) reports that approximately 80 million people globally need a wheelchair~\cite{puli2024global}, including around 5.5 million in the United States, a number that has doubled over the past decade~\cite{USDOT2024}. Because the human upper limb is essential for ADLs and independence~\cite{poirier2023efficient}, WMRAs are increasingly being explored in assistive robotics to restore functionality. However, while wheelchairs and robotic arms have been studied extensively, their integration is underexplored. Complex control interfaces and frequent mode switching between two devices~\cite{chung2017performance, chung2024robotic} create cognitive overhead and pose barriers to practical deployment.

\begin{figure}[t]
    \centering
    \includegraphics[width=0.9\linewidth]{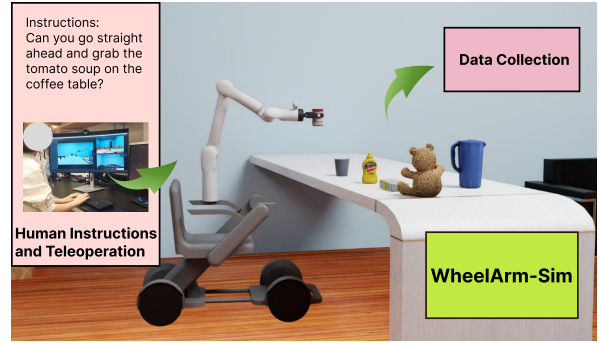}
    \caption{WheelArm-Sim features a human-in-the-loop workflow where teleoperation guides task execution in a physics-based simulator, with real-time data collection to record human instructions, RGB-D images, and robot data.}
    \label{fig:cover_short} 
\end{figure}

Recent advances in machine learning and AI have enabled intuitive control methods to fuse different modalities into robot actions, which has potential for the WheelArm intuition control. Vision-Language Action Models (VLAs), such as RT series~\cite{brohan2022rt, brohan2023rt} and $\pi_0$~\cite{black2024pi_0}, offer an end-to-end method to interpret human instructions and translate them into robot actions using vision and language inputs to perform manipulation and navigation tasks. Despite VLA's promise, data scarcity remains a fundamental challenge~\cite{firoozi2023foundation}. Collecting large-scale, diverse, and safe real-world data is time-consuming and resource-intensive, especially for researchers who lack access to a physical WheelArm but still wish to leverage its data.

High-fidelity simulators enable scalable, safe, and cost-effective solutions for VLA data collection. Current simulation platforms~\cite{gan2020threedworld, li2023behavior} typically focus on general robotic tasks and neglect users' needs who rely on assistive systems that combine wheelchairs and robotic arms. Furthermore, most simulators lack integrated support for systems combining wheelchairs and robotic arms.

Our research aims to build a physics-based simulator for WheelArm data collection to facilitate its integrated control (\textbf{Figure~\ref{fig:cover_short}}). To address these limitations, we introduce WheelArm-Sim, an indoor simulation environment built in Isaac Sim. WheelArm features a Kinova Gen3 Robotic Arm mounted on a Whill Model CR2 wheelchair within a realistic home environment designed to simulate ADLs. We further present a small-scale multimodal dataset collected using this simulator, including human instructions, RGB-D images from the arm wrist camera and the wheelchair-mounted camera, Inertial Measurement Unit (IMU) data, whole-body joint states, and relative Cartesian poses and velocities. Finally, we use a simple baseline model to predict WheelArm actions based on prior actions. This demonstrates the simulator's potential to support data collection for VLA development on WheelArm. 

In summary, the contributions of this paper are: (i) a physics-based WheelArm simulator in Isaac Sim using models of the Kinova Gen3 arm, Whill Model CR2 wheelchair, and an indoor ADL environment, (ii) a Robot Operating System (ROS2)-based data collection workflow with Graphical User Interface (GUI) and keyboard teleoperation, (iii) a small three-modality dataset includes two activities, 13 tasks, and 232 trajectories, and (iv) an analysis of a baseline model that takes time-ordered past data as input to predict the robot's next actions.
\vspace{-1em}

\section{Related Work}
\label{sec:related work}\subsection{Assistive Robotics}
WMRAs have progressed significantly in design, control, and user interaction. For design, affordable robotic arm prototypes provides cost-effective solutions for robotic arm control~\cite{leone2024design}. Additionally, shared control has progressed through diverse inputs, which interpret human intentions via voice commands ~\cite{babu2021review}, electroencephalogram (EEG)~\cite{sultan2025transnn} and electromyography (EMG) signals~\cite{cheng2022robotic}, as well as eye tracking~\cite{cio2019proof, fischer2024scoping}. This work is complemented by research in autonomous wheelchair control ~\cite{yanco2006wheelesley, xu2025conav} which focuses on navigation and Simultaneous Localization and Mapping (SLAM) to enhance user experience.
\vspace{-0.4em}

\subsection{Simulation}
\vspace{-0.2em}
Simulators play a crucial role in algorithm testing, reinforcement learning, and synthetic data collection. Platforms like Habitat~\cite{savva2019habitat,szot2021habitat,puig2023habitat} and ALFRED~\cite{shridhar2020alfred} have advanced in path planning and task decomposition, but lack detailed physics in object-gripper interactions. The iGibson series~\cite {shen2021igibson,li2021igibson,li2024behavior}, ManiSkill-Hub~\cite{shukla2024maniskill}, and VirtualHome~\cite{puig2018virtualhome} focus on physics-based domestic tasks but do not address the needs of users with physical limitations relying on wheelchairs and WMRAs. RCareWorld~\cite{ye2022rcare} addresses this gap by simulating care environments with physics-enabled avatars and caregivers. However, since it is not designed for the integrated WheelArm control, its settings and configuration do not support multimodal sensor data, such as movable RGB-D cameras and wheel actuator positions, required by an AI-based control model. Our simulator stands out by being specifically designed for wheelchair–arm systems and by generating synchronized multimodal data to support joint policy learning. \vspace{-0.4em}

\subsection{Datasets}
\vspace{-0.2em}
Various robots have collected diverse datasets across a range of settings. OpenX-Embodiment (OXE)~\cite{robotics_transformer_x} comprises diverse morphological robot datasets, covering a broad spectrum of skills such as picking, moving, and pushing. For example, D3Fields~\cite{wang2023d} contributes 3D rearrangement data collected using a Kinova Gen3 robot. However, few existing datasets can be used for training integrated wheelchair and robotic arm models since few satisfy the requirements of containing assistive tasks, multimodal data, and similar operational settings simultaneously.

To our knowledge, our work introduces the first physics-based simulation for integrated WheelArm multimodal data collection. WheelArm-Sim fills the current research gap by combining assistive tasks in an indoor environment and establishing a comprehensive multimodal sensor data collection pipeline for training a potential VLA model for integrated control. \vspace{-0.5em}

\section{Activities and Tasks}
\label{sec:Activities and Tasks}This section outlines selected activities and tasks for data collection, focusing on essential ADLs for independent living. Based on the WHO's International Classification of Functioning, Disability, and Health (ICF), these tasks fall within the "Activities and Participation" domain, specifically in mobility, self-care, and domestic life. Key movements include lifting, grasping, reaching, and navigating the home. The tasks are classified into two categories: (i) organization activities, which involve arranging and positioning objects, and (ii) serving yourself activities, which focus on direct interactions with objects for self-care, like picking up food or drinks. \textbf{Table \ref{tab:activity_comparison}} details the activities and corresponding tasks selected. 

Instead of limiting the data collection to traditional pick-and-place tasks, wheelchair navigation data is integrated to make the dataset unique in covering manipulation and movement dynamics. Additionally, we include the task of opening a drawer because it is a significant robot task in cooking and organizing, as well as in the motion planning field. Similarly, selecting food or drink is an essential task in ADLs. Since the simulation does not include a human model in this version, tasks like feeding are excluded. A varied set of objects designed for different types of interaction is detailed in \textbf{Table \ref{table:objects}}. 

\vspace{-0.5em}

\begin{table}[h]
\centering
\caption{Adapted Activity Descriptions Chosen based on ICF. These tasks ensure the dataset captures manipulation and navigation for future intuitive WheelArm control developments.}
\label{tab:activity_comparison}
\begin{tabular}{|p{2.1cm}|p{5.25cm}|}
\hline
\textbf{Activity} & \textbf{Tasks} \\
\hline
\textbf{Organization Activities}  
& 
- Navigate to a target location and pick up kitchenware. \newline
- Navigate to a target location, pick up objects such as a knife, and place them in a designated area. \newline
- Navigate to a drawer and open it. \\
\hline
\textbf{Serving Yourself Activities}  
& 
- Navigate to a designated location and select food or drink. \\
\hline
\end{tabular} 
\end{table}

\vspace{-0.9em}


\begin{table}[h]
\centering
\caption{Objects in two Activities. A diverse selection of rigid and deformable objects ensures a comprehensive dataset.}
\label{table:objects}
\begin{tabular}{|p{2.1cm}|p{5.25cm}|}
\hline
\textbf{Activity} & \textbf{Objects}\\
\hline
\textbf{Organization Activities}  
& 
Knife, Grey Cup, Blue Mug, Silver Bottle, Glass Bottle, Bowl, Teddy Bear, Drawer \\
\hline
\textbf{Serving Yourself Activities}  
& 
Mustard, Crackers, Tomato Soup, Meat Can, Coke \\
\hline
\end{tabular} \vspace{-2em}
\end{table}

\section{WHEELARM SIMULATOR}
\label{sec:simulator}\subsection{Simulation Scenario}
\begin{figure}[!ht]
    \centering
    \includegraphics[width=0.45\textwidth]{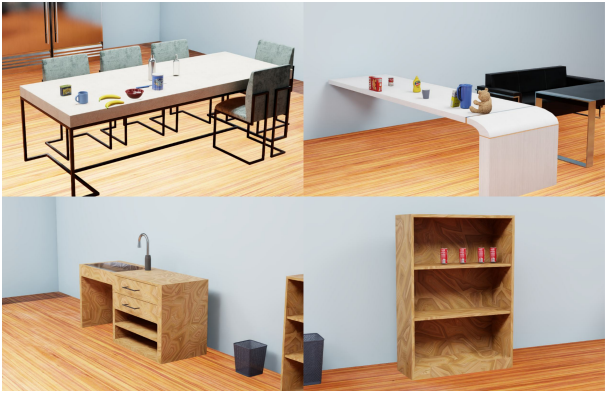}
    \caption{Four areas for tasks in the WheelArm-Sim environment.}
    \label{fig:four_areas} 
\end{figure}

The WheelArm-Sim environment is created using Isaac Sim to build a realistic and physics-based living room for the WheelArm. The environment includes four task-specific areas: a kitchen table for pick-and-place, a coffee table for general object interaction, a kitchen workstation for drawer opening, and a shelf for Coke selection. \textbf{Figure \ref{fig:four_areas}} depicts the layout of the WheelArm-Sim environment. 

\begin{figure}[!ht]
    \centering
    \includegraphics[width=0.4\textwidth]{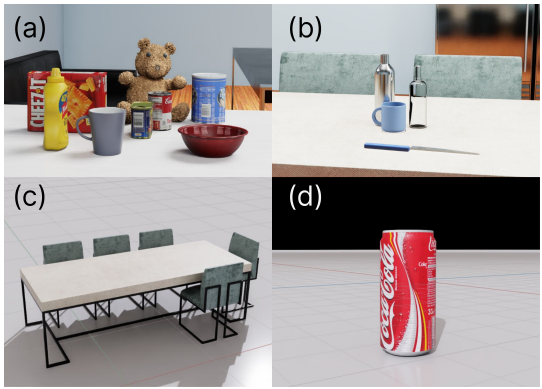}
    \caption{(a) Isaac Sim built-in objects, (b) self-created objects, and (c,d) online objects. Our dataset includes diverse object types, including deformable materials. }
    \label{fig:Objects} 
\end{figure}

To ensure a diverse dataset, the simulation incorporates objects from various sources, including Isaac Sim built-in objects like tomato soup and crackers, self-designed objects, and online objects (\textbf{Figure \ref{fig:Objects}}). The teddy bear, which is a deformable object, properties like Young's Modulus, dynamic friction, elasticity damping, and Poisson's Ratio were set to 7000 Pa, 25, 0.005, and 0.4, respectively, to mimic the real grasping reaction. We created custom objects (\textbf{Figure~\ref{fig:Objects} (b)}), including a knife, a glass bottle, and a silver bottle, using Onshape, a web-based CAD platform compatible with Isaac Sim. All assemblies in Onshape consist of nested parts and subassemblies for efficient object management.



\subsection{WheelArm Simulation Model}
The WheelArm simulation model combines the Kinova Gen3 robotic arm with the Whill Model CR2 wheelchair in Isaac Sim for coordinated manipulation and navigation tasks. (see the WheelArm simulation model in \textbf{Figure \ref{fig:cover_short}}). The Kinova Gen3 arm is controlled using inverse kinematics via the screw-theory-based Newton-Raphson method~\cite{Babaiasl2024}. For navigation, we use Isaac Sim's Action Graph to create ROS2 nodes for differential driving.
\vspace{-0.2em}
\subsection{Computation Device}
We used a Dell Precision 3660 with an Intel Core i7 processor and an NVIDIA A4000 GPU for both simulation and data collection. The model training was completed on a high-performance cluster (HPC) featuring A100 GPUs. \vspace{-0.8em}

\section{DATA COLLECTION}
\label{sec:collection}
This section presents the ROS2-based, human-in-the-loop data collection workflow and the inverse kinematics approach for controlling the WheelArm in simulation.\vspace{-0.5em}

\subsection{Data Collection Workflow}
\begin{figure}[htbp]
    \centering \vspace{-0.8em}
    \includegraphics[width=0.45\textwidth]{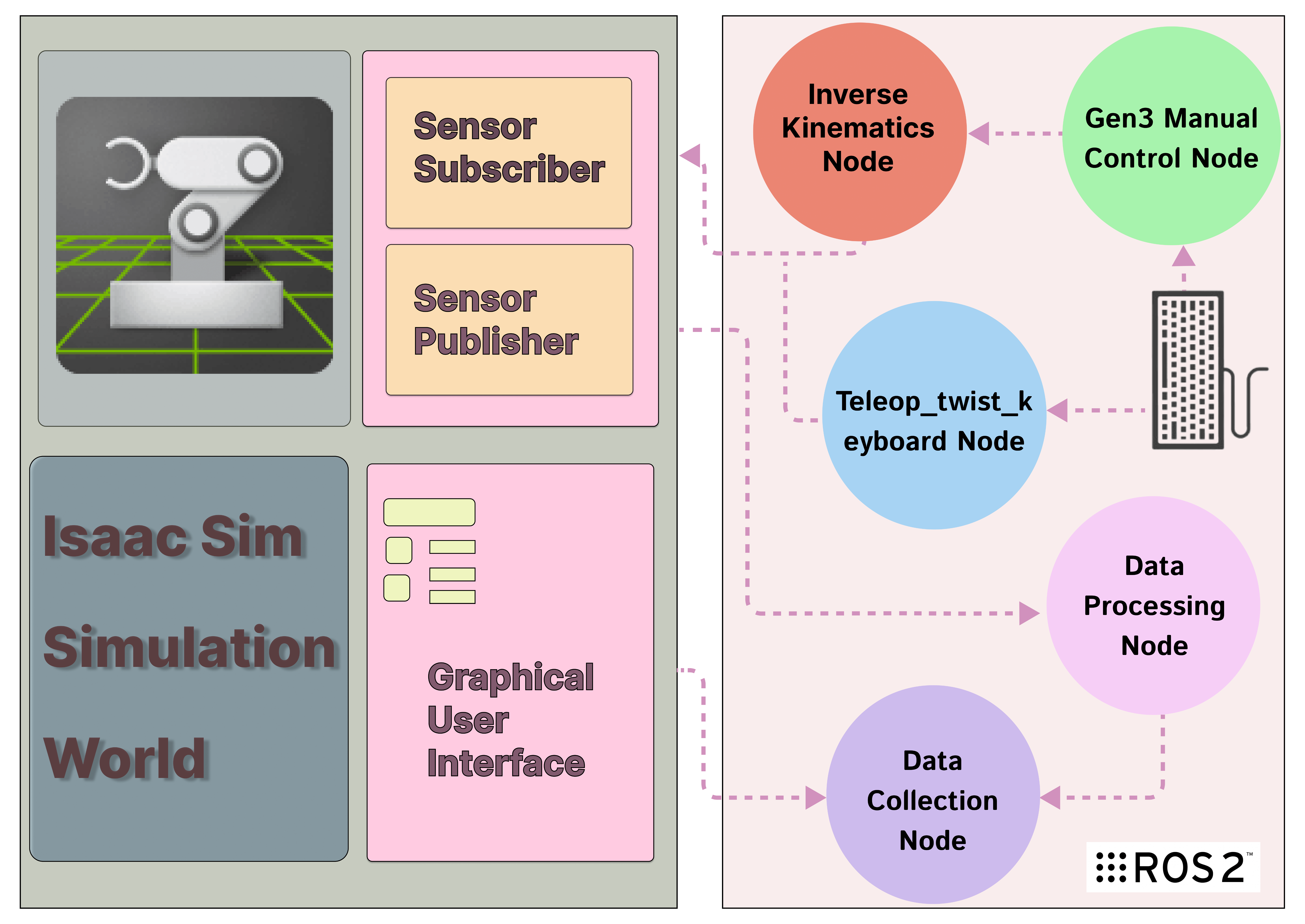}
    \caption{ROS2-based Data Collection Workflow integrating user teleoperation, robotic control, and sensor data processing. This workflow enables efficient multimodal data collection for the WheelArms.}
    \label{fig:data_collection_workflow} 
\end{figure}

The data collection workflow is developed on ROS2 and incorporates user teleoperation, robotic control, and a data-saving GUI within Isaac Sim. This setup allows researchers to collect data efficiently, as shown in \textbf{Figure~\ref{fig:data_collection_workflow}}. 
Gen3 Manual Control Node, Inverse Kinematics Node, and Teleop\_twist\_keyboard Node link the user inputs and robot execution in Isaac Sim. The keyboard input publishes the robotic arm's target goals, controls the gripper opening and closure, and manages wheelchair velocity. The Gen3 Manual Control node adjusts the target poses of the end-effector and gripper actuators by 0.025 meters or 0.05 radians per keyboard click. The Inverse Kinematics Node computes the corresponding joint positions using the screw theory-based Newton-Raphson method \cite{Babaiasl2024} and transmits the calculated values to the articulation node in Isaac Sim for execution. The Differential Driving Node from Isaac Sim listens to the velocity commands from the Teleop\_twist\_keyboard Node in the forward, backward, left, and right directions.

Sensor publishers transmit robot joint states, images, joint coordinate transformations, and Inertial Measurement Unit (IMU) data to the Data Processing Node which organizes data by corresponding timestamps and topic names. The GUI, an Isaac Sim extension, is created to name data collection files, input human instructions, and trigger the start and end times for data saving.

The data saving system uses a Data Collection Node to capture and store RGB-D images, human instructions, and robot action data in a compressed HDF5 format.

\subsection{Inverse Kinematics Method}
We use the screw-theory-based Newton-Raphson method \cite{Babaiasl2024} to map the desired end effector Cartesian pose into joint actuator positions in the simulator. The pseudocode for the inverse kinematics is presented in \textbf{Algorithm~\ref{alg:newton_raphson}}.

\begin{algorithm}
\caption{Inverse Kinematics using Screw Theory}
\begin{algorithmic}[1]
\Require $T_{sd}$ (desired end-effector transformation), $q_0 \in \mathbb{R}^n$ (initial joint guess), tolerance $\epsilon$
\State Initialize $i \gets 0$.
\While{not converged}
    \State Compute error twist $\mathcal{V}_b$: $[\mathcal{V}_b] = \log(T_{sb}^{-1}(q_i) T_{sd})$
    \If{$\|\mathcal{V}_b\| < \epsilon$} \Comment{Check convergence}
        \State \textbf{break}
    \EndIf
    \State Compute joint update: $q_{i+1} = q_i + J_b^{\dagger}(q_i) \mathcal{V}_b$
    \State $i \gets i + 1$
\EndWhile
\State \Return $q_i$ (final joint configuration)
\end{algorithmic} 
\label{alg:newton_raphson} 
\end{algorithm} 
\vspace{-1.2em}

\section{DATASET STRUCTURE}
\label{sec:dataset}The dataset contains 20 recorded trajectories per object for picking (excluding mustard) and opening drawers. In the pick-and-place task, there are 10 trajectories for a teddy bear, 9 for a knife, and 9 for Coke selection. Each experiment includes human instructions, RGB-D images, and robot joint-related data.

\noindent
\textbf{Instructions.} The dataset includes human language instructions for tasks, like "Turn right, go to the coffee table, and pick up the mustard," which guide navigation and interactions between the robot and objects.

\noindent
\textbf{RGB-D Images.} WheelArm-Sim gathers images from mobile cameras mounted on the WheelArm to perceive the environment from the robot’s perspective for object detection and localization.

\noindent
\textbf{Robot Joint-related Data.} For the wheelchair, we capture its Cartesian pose, velocities, IMU data, and the positions and velocities of all four wheels. Additionally, we record each joint’s Cartesian pose, velocities, and joint states (positions and velocities) for the robotic arm. \vspace{-0.5em}


\section{BASELINE}
\label{sec:baseline}We designed a baseline and applied it to the mastard-picking task to explore the potential of using this three-modality dataset for action prediction. \vspace{-0.7em}

\subsection{Baseline Structure}
The baseline model follows a multimodal sequence learning approach to predict the next-step actions of the WheelArm. 
The architecture employs an LSTM-based sequential model that processes multiple input modalities and fuses them into a shared representation. Each modality is encoded using specialized feature extractors. We use a pre-trained ResNet18 architecture for RGB images encoding by removing its final fully connected layer. Depth features are extracted using a simple CNN with two convolutional layers. Additionally, human instructions are encoded with a pre-trained Word2Vec model, which is fine-tuned during training. A fully connected layer projects the text embeddings into a learned feature space. Joint-related data are processed using a linear network with a ReLU activation function and a normalization layer, while timestamps are encoded into embeddings with a linear layer to express time information and capture joint movement speed. These concatenated features are passed through a Multilayer Perceptron (MLP) fusion module to reduce dimensionality before entering the LSTM network. The final prediction layer outputs the next robotic pose, including the Cartesian positions (x,y,z) and orientation (quaternion x,y,z,w) of the robotic arm and wheelchair, and the gripper actuator joint positions (in radians). \textbf{Table~\ref{table:baseline_hyperparameters}} summarizes the baseline model’s architecture and hyperparameters.


\vspace{-0.6em}
\begin{table}[h]
\centering
\caption{Baseline Model Hyperparameters and Network Layers.}
\label{table:baseline_hyperparameters}
\resizebox{\columnwidth}{!}{ 
\begin{tabular}{|l|c|}
\hline
\textbf{Parameter}           & \textbf{Value} \\
\hline
LSTM Hidden Dimension       & 128 \\
LSTM Layers                 & 1 \\
Instruction Embedding Size  & 300 (Word2Vec) \\
MLP Fusion Output Dimension & 128 \\
Depth CNN Output Size       & 64 \\
Learning Rate               & \(1\times10^{-4}\) \\
Weight Decay                & \(1\times10^{-4}\) \\
Batch Size                  & 16 \\
Sequence Length             & 20 \\
Optimizer                   & Adam \\
Loss Function               & MSE \\
LR Scheduler                & StepLR (factor 0.1, step size 5) \\
Gradient Clipping           & Max norm 1.0 \\
Maximum Epochs              & 100 \\
Train-Validation Split      & 80\%-20\% \\
Early Stopping Patience     & 3 \\
\hline
\end{tabular}
} 
\end{table}
\vspace{-1.5em}

\subsection{Training and Testing Data}

In the mustard-picking task, 24 experiments were conducted, 20 for training and validation, and 4 for testing. Specifically, experiments 1 to 16 were for training, and 17 to 20 were for validation. The model was trained using a sequence length of 15 timesteps, with an 80-20 split for the training and validation data.
\vspace{-0.5em}

\section{RESULTS}
\label{sec:Results}\vspace{-0.5em} 
This section visualizes the dataset distribution and outlines the data processing methodology. We present quantitative performance metrics for the training and validation datasets and an example of model predictions from the test dataset. These results allow us to evaluate the performance of using data collected from the simulator in a baseline using an LSTM-based architecture.

\vspace{-0.7em}
\subsection{Data Analysis}
\subsubsection{Dataset Composition}
\begin{figure}[!ht]
    \centering \vspace{-0.5em} 
    \includegraphics[width=0.4\textwidth]{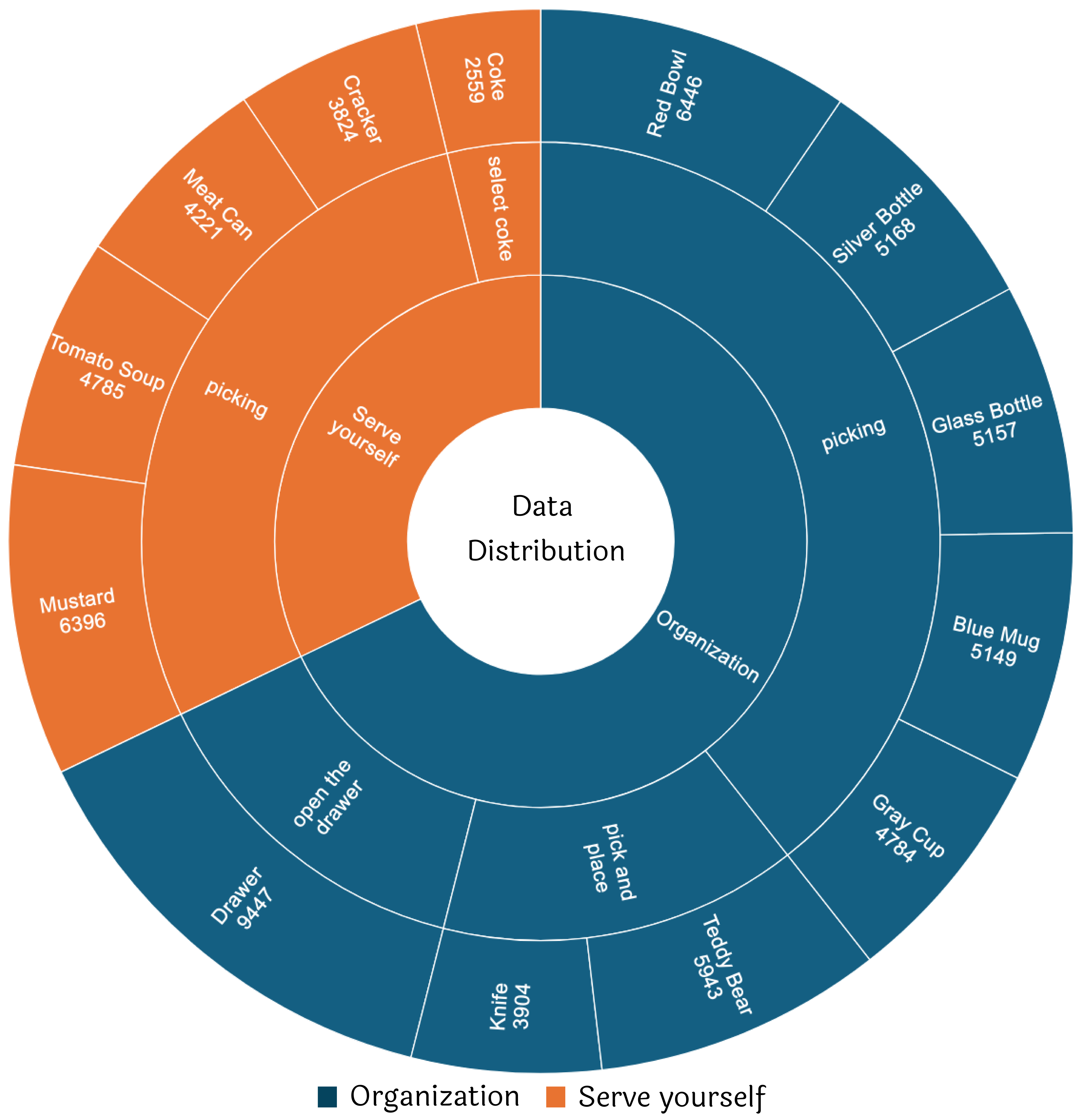}
    \caption{Dataset Distribution}
    \label{fig:data_distribution}
\end{figure}

The dataset comprises two activities consisting of 13 tasks, 232 trajectories, and a total of 67,783 samples. The Organization activity contains 45,998 samples, while the Serve Yourself activity has 21,785 samples. For the Organization, picking, pick-and-place, and opening the drawer have 26,704, 9,847, and 9,447 samples, respectively. \textbf{Figure~\ref{fig:data_distribution}} illustrates the dataset's distribution. 


\subsubsection{Data processing}
\begin{figure}[htbp]
    \centering
    \includegraphics[width=0.45\textwidth]{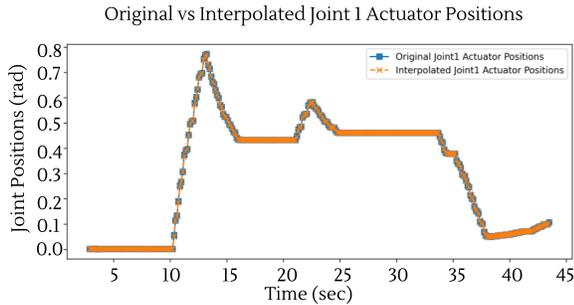}
    \caption{Original vs. Interpolated Joint 1 Actuator Positions.}
    \label{fig:joint_pose_sync}\vspace{-1em}
\end{figure}

\begin{figure}[!h]
    \centering
    \includegraphics[width=0.45\textwidth]{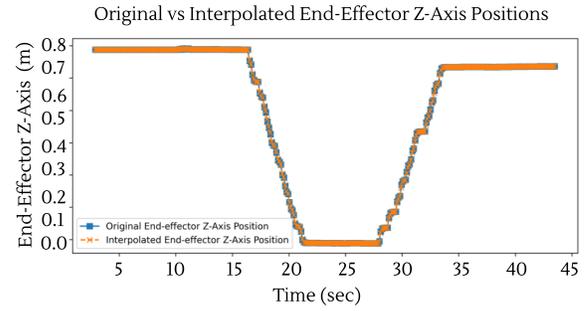}
    \caption{Original vs. Interpolated End Effector Z-Axis Positions.}
    \label{fig:end_effector_sync} \vspace{-1em}
\end{figure}

\begin{figure}[!ht]
    \centering
    \includegraphics[width=0.45\textwidth]{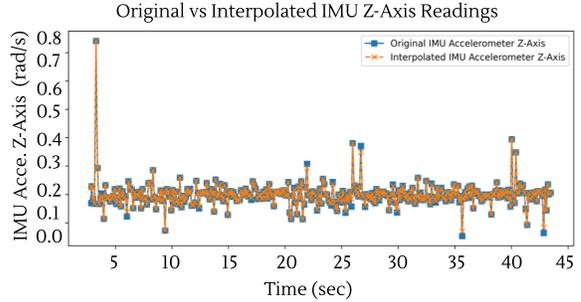}
    \caption{Original vs. Interpolated IMU Z-Axis Reading.}
    \label{fig:imu_sync} 
\end{figure}

The Data Processing Node (Figure:~\ref{fig:data_collection_workflow}) collects data asynchronously, resulting in different timestamps for Cartesian robot poses, IMU data, images, and joint states. To unify these streams, we chose image timestamps as the reference frame and used linear interpolation to align the robot pose, IMU, and joint states.  A comparison of the interpolated and original data for joint 1 is shown here (the same method applies to other joints). The actuator positions, end effector positions, and IMU z-axis reading for joint 1 represent the joint states, Cartesian robot poses, and IMU data, respectively. \textbf{Figures~\ref{fig:joint_pose_sync},~\ref{fig:end_effector_sync}, and~\ref{fig:imu_sync}} illustrate the alignment of original data across different timestamps with the interpolated results. The interpolated data (orange) smoothly tracks the original data (blue), confirming effective shape preservation at the reference times.

\vspace{-1em} 

\subsection{Baseline Analysis}
After quantitative evaluation, we visually examine trajectory trends in the test dataset. Unlike existing VLA models~\cite{brohan2022rt,brohan2023rt,kim2024openvla}, which require extensive datasets, our focus is on understanding the performance of our simulator-derived tasks. We aim to determine which prediction aspects succeed and which do not, and whether action prediction helps the robot approach the target. This analysis demonstrates the potential of using simulated data for unified WheelArm control. A prediction is deemed successful if it effectively captures overall trends.

\subsubsection{Baseline Quantitative Evaluation}
After training the baseline model with 16 experiments from the mustard-picking task, we assessed its predictive performance using MAE and MSE metrics. MSE highlights larger errors (sensitive to outliers), while MAE provides a balanced view of average absolute errors. Results (\textbf{Tables~\ref{tab:train_combined_metrics} and~\ref{tab:val_combined_metrics}}) show error variability across components. The large range in Wheelchair X-Positions for both training and validation indicates high prediction variability, though this is influenced by the large range of ground truth values. Conversely, the small range in Wheelchair Y-Orientations results in low MSE and MAE, suggesting stability that is partly due to the low inherent ground truth variability. Overall, the model is most sensitive to errors in the wheelchair's x-position and least sensitive to its y-orientation (quaternion). \vspace{-0.5em}

\begin{table}[h]
    \centering
    \caption{Training Dataset Combined MSE/MAE Metrics. EE: End Effector; W: Wheelchair; GR: Gripper Right; GL: Gripper Left. This table displays the minimum and maximum values for MSE and MAE among predictions. The wheelchair's x-axis position is the most sensitive feature, while its y-orientation is the least sensitive.} 
    \scalebox{0.8}{  
    \begin{tabular}{lcc}
        \toprule
        \textbf{Predictions} & \textbf{Min MSE/MAE Value} & \textbf{Max MSE/MAE Value} \\
        \midrule
        EE X-Positions (m) & 9.5683e-04/2.6495e-02 & 2.2572e-02/1.3904e-01 \\
        EE Y-Positions (m) & 5.9490e-04/1.9270e-02 & 2.0771e-02/1.1210e-01 \\
        EE Z-Positions (m) & 7.4201e-04/2.0612e-02 & 2.3753e-02/1.4358e-01 \\
        EE X-Orientations  & 4.7718e-03/5.7165e-02 & 3.3369e-01/5.0193e-01 \\
        EE Y-Orientations & 3.3648e-03/4.6756e-02 & 2.9342e-01/4.6361e-01 \\
        EE Z-Orientations & 4.3206e-03/5.2389e-02 & 2.7580e-01/4.5399e-01 \\
        EE W-Orientations & 3.3736e-03/5.0158e-02 & 2.9016e-01/4.6493e-01 \\
        W X-Positions (m) & 1.4072e-02/9.1089e-02 & 1.7957e+00/9.3832e-01 \\
        W Y-Positions (m) & 4.0096e-04/1.5062e-02 & 2.0463e-02/1.2637e-01 \\
        W Z-Positions (m) & 1.7875e-13/3.4645e-07 & 2.9761e-12/1.3821e-06 \\
        W X-Orientations & 3.6380e-14/1.4761e-07 & 5.2281e-13/5.7871e-07 \\
        W Y-Orientations & 2.2705e-14/1.2243e-07 & 2.7310e-13/4.6102e-07 \\
        W Z-Orientations & 1.3052e-05/2.9225e-03 & 4.5435e-04/1.9752e-02 \\
        W W-Orientations & 3.2860e-09/4.7412e-05 & 4.5611e-08/1.8716e-04 \\
        GL Actuator Positions (rad) & 1.1893e-03/2.6235e-02 & 4.3060e-02/1.9901e-01 \\
        GR Actuator Positions (rad) & 6.1458e-04/1.8515e-02 & 3.5521e-02/1.8132e-01 \\
        \bottomrule
    \end{tabular}
    }
    \label{tab:train_combined_metrics} 
\end{table}

\begin{table}[h]
    \centering
    \caption{Validation Dataset Combined MSE/MAE Metrics: the wheelchair's x-axis position is the most sensitive feature, while the y-axis orientation is the least sensitive.}
    \label{tab:val_combined_metrics}
    \scalebox{0.8}{  
    \begin{tabular}{lcc}
        \toprule
        \textbf{Predictions} & \textbf{Min MSE/MAE Value} & \textbf{Max MSE/MAE Value} \\
        \midrule
        EE X-Positions (m) & 1.1566e-04/8.2090e-03 & 1.5735e-02/1.2520e-01 \\
        EE Y-Positions (m) & 1.1199e-04/9.1092e-03 & 1.9473e-02/1.3905e-01 \\
        EE Z-Positions (m) & 1.3488e-04/8.3293e-03 & 3.3810e-02/1.8191e-01 \\
        EE X-Orientations & 3.6350e-03/5.1540e-02 & 4.0448e-01/5.9895e-01 \\
        EE Y-Orientations & 4.1162e-03/5.8829e-02 & 4.8562e-01/6.9581e-01 \\
        EE Z-Orientations & 3.3662e-03/5.0593e-02 & 4.2624e-01/6.4680e-01 \\
        EE W-Orientations & 4.1796e-03/4.8868e-02 & 4.0180e-01/6.3323e-01 \\
        W X-Positions (m) & 2.3338e-04/1.2938e-02 & 1.3200e+00/9.0051e-01 \\
        W Y-Positions (m) & 1.6643e-05/3.2871e-03 & 8.8342e-03/9.3283e-02 \\
        W Z-Positions (m) & 3.1086e-14/1.3039e-07 & 1.2701e-12/1.0505e-06 \\
        W X-Orientations & 7.6967e-14/1.8806e-07 & 3.7358e-13/5.1148e-07 \\
        W Y-Orientations & 4.6379e-14/1.6407e-07 & 2.4714e-13/3.8500e-07 \\
        W Z-Orientations & 2.3387e-07/4.2310e-04 & 1.9702e-04/1.3917e-02 \\
        W W-Orientations & 3.6345e-10/1.5397e-05 & 6.7195e-08/2.3710e-04 \\
        GL Actuator Positions (rad) & 8.5456e-05/7.7381e-03 & 6.5904e-02/2.3717e-01 \\
        GR Actuator Positions (rad) & 4.7816e-05/5.8004e-03 & 5.8092e-02/2.3360e-01 \\
        \bottomrule
    \end{tabular}
    }\vspace{-2.3em} 
\end{table}  

\subsubsection{Baseline Qualitative Evaluation}
\begin{figure}[h]
    \centering
    \includegraphics[width=0.35\textwidth]{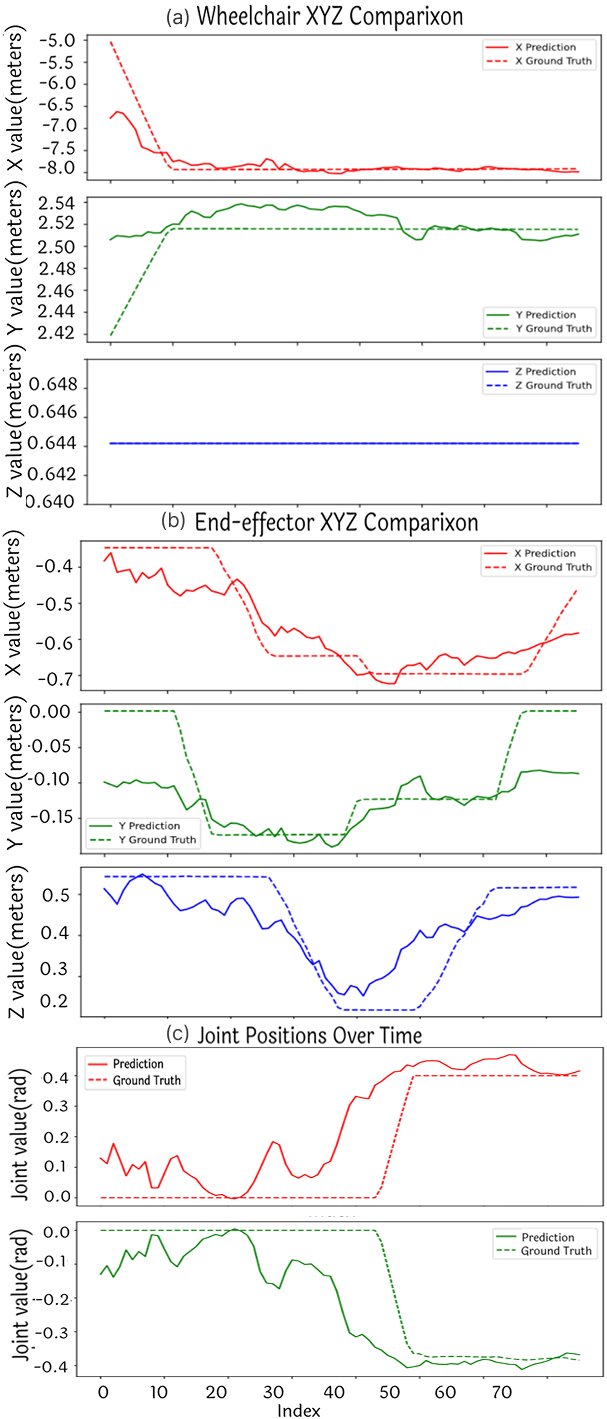}
    \caption{Predicted vs. Ground Truth: (a) Captures x-axis sliding-window indexing and relative y-values (missing the initial rise); z-axis matches closely (overlapping curves). (b) Captures X trends and step-like Y patterns, with timing offsets in Z. (c) Gripper closing is captured with a slight timing offset.}
    \label{fig:visualization}
\end{figure}

To visually assess alignment, we compare predicted and ground-truth position trajectories for the wheelchair, end-effector, and gripper actuators (\textbf{Figures~\ref{fig:visualization} (a-c)}). Results show the model performs better on simpler trajectories and worse on more complex ones.

In wheelchair movement prediction (\textbf{Figure~\ref{fig:visualization} (a)}), the model generally captures the average movement in x, y, and z. While it misses the initial upward trend along the y-axis, it accurately predicts the constant z-axis height (overlapping the ground truth). The overall success in predicting this straightforward turn-right-then-forward trajectory demonstrates the potential for predicting simple tasks like wheelchair movement.

End-effector position predictions (\textbf{Figure~\ref{fig:visualization} (b)}) present challenges, as the model partially captures trends but misses details. Along the x-axis, the model predicts the overall downward/upward trend but misses short-term decreases and misjudges the timing of the upword transition. For the y-axis, it captures the step-like trend but fails to match exact start/end values, which is acceptable since our focus is on trend-following. On the z-axis, the model successfully predicts the decreasing and subsequent increasing trend but initiates the upward movement early.

The model's performance on the gripper is mixed (\textbf{Figure~\ref{fig:visualization} (c)}) . It successfully predicts the closure by predicting the simultaneous changes in the left and right actuators, but the timing is slightly off, starting earlier than the ground truth.

Overall, the baseline model demonstrates promise in predicting simple position trajectories (wheelchair navigation) while requiring improvement in handling complex end-effector orientation and timing accuracy.
\vspace{-0.8em}

\section{CONCLUSIONS and FUTURE WORK}
\label{sec:result}
\vspace{-0.5em}
This work introduces a physics-based indoor simulator for collecting WheelArm multimodal data in human-in-the-loop settings, resulting in a dataset of 13 tasks and 67,783 samples. An LSTM baseline shows promise for simple action prediction but struggles with complex tasks due to limited data. WheelArm-Sim supports integrated control, but limitations persist: the dataset is small, restricting model generalization; the baseline could be improved with a VLA approach; user experience needs improvement (upgrading from keyboard to VR headsets); and the sim-to-real gap remains unaddressed. Future work will focus on overcoming these limitations.
\vspace{-1em} 






\section*{ACKNOWLEDGMENT}
\vspace{-0.5em}
This work was partially supported by NSF Awards \# 2201536 and \#2430236. 



\bibliographystyle{IEEEtran}
\bibliography{reference}

\end{document}